\documentclass[runningheads]{llncs}
\usepackage[T1]{fontenc}
\usepackage{graphicx}
\usepackage{graphicx}
\usepackage{amsmath,amssymb}

\begin{document}

\title{Agentic Generative AI for Media Content Discovery at the National Football League}

\titlerunning{Agentic Generative AI for Media Content Discovery at the NFL}
%
\author{Henry Wang\inst{1}\and
 Md Sirajus Salekin\inst{1}\and
 Jake Lee\inst{1}\and 
 Ross Claytor\inst{1}\and 
 Shinan Zhang\inst{1}\and
 Michael Chi\inst{2}}
\authorrunning{H. Wang et al.}
%
\institute{Amazon Web Services, Seattle, USA \and
National Football League, New York, USA
}
\maketitle              
\begin{abstract}
Generative AI has unlocked new possibilities in content discovery and management. Through collaboration with the National Football League (NFL), we demonstrate how a generative-AI based workflow allows media researchers and analysts to query relevant historical plays using natural language, rather than using traditional filter and click-based interfaces. The agentic workflow takes a user query in natural language as an input, dissects the query into different elements, and then translates these elements into the underlying database query language. The accuracy and latency of retrieval are further improved through carefully designed semantic caching. The solution performs with over 95-percent accuracy and reduces the average time of finding relevant videos from 10 minutes to 30 seconds, significantly increasing the NFL’s operational efficiency and allowing users to focus more on producing creative content and engaging storylines.

\end{abstract}
%
%
%

\section{Introduction}

The demand for sports content grows every year, as shown by the increasing sizes of sports leagues' multi-year media rights deals valued at billions of US dollars \cite{elberse2022nfl,guardian2021nfl}. The National Football League (NFL)\footnote{\url{https://www.nfl.com/}}, one of the most popular professional sports leagues in the world, produces and distributes content all year-round to meet this demand and to engage its 184 million domestic and 100 million international fans. The NFL  manages petabytes of content, primarily videos of game footage, and distributes it for use across broadcasts, TV segments, digital platforms (sites, apps), and social media. In a sports entertainment landscape that is becoming increasingly more digital, producing captivating content remains at the core of what the NFL delivers to fans around the world. However, it has become increasingly difficult for media teams to find and manage the right content in the NFL’s ever-growing library of digital media. 

The NFL partnered with our team to develop an agentic solution - powered by generative AI - that enables media and production teams to efficiently search for game videos using natural language queries\footnote{\url{https://www.wired.com/sponsored/story/will-the-nfls-push-into-genai-transform-how-we-see-sports/}}. The solution enables users to spend more time on the creative aspects of their roles to create more engaging content for fans.

\section{Related Works}
The advent of generative AI has revolutionized traditional content retrieval paradigms, transforming how we interact with and extract information from vast data repositories. With the power of Large Language Models (LLMs) \cite{zhu2023large}, we can now perform text-to-SQL and Retrieval-Augmented Generation (RAG) operations that convert natural language queries into corresponding intermediate representations of the SQL queries, vector embeddings, or other structured formats, thus, enabling more intuitive and effective content retrieval. Apart from the general in-context learning capability of LLMs, recent text-to-SQL approaches can solve different complex text-to-SQL tasks using chain-of-thoughts \cite{tai2023exploring}, few-shot strategy \cite{xie2025opensearch}, decomposing steps \cite{xie2024decomposition}, self correction \cite{yuan2025cogsql}, and multi-turn steps \cite{zhang2024coe}. On the other hand, RAG \cite{fan2024survey} frameworks can retrieve relevant context for any particular query. Both text-to-SQL and RAG systems can retrieve domain-specific content based on SQL or vector embeddings.

In sports, media content retrieval is a common task when preparing highlights or writing digital posts. Recently, LLMs have shown their potential in transforming how sports organizations manage, analyze, retrieve, and discover content within their vast media archives. LLMs are being used for applications including: expert commentary generation \cite{li2025multi,cook2024llm,ge2024scbench,sarfati2023generating}, action spotting \cite{shin2025soccer}, sports understanding \cite{xia2024sportu,yang2024sports,jiang2025domain,schilling2024querying,lee2024sportify,zhang2025chatmatch}, and video assistant refereeing systems \cite{held2024x}. Natural language queries are making sports content more accessible to analysts, fans, and broadcasters. Related work in soccer (football) has shown rather than relying on structured search parameters or specific keywords, users can now query and retrieve soccer information using RAG \cite{li2025multi,strand2024soccerrag} or GraphRAG \cite{sepasdar2024soccer}.

While LLMs introduce the capability to perform natural language queries and to retrieve relevant content, the extent to which LLMs can be utilized to understand and streamline complex data and media platforms (such as the NFL’s Next Gen Stats platform) is yet to be explored. Specifically, it would be worthwhile to explore whether LLMs and natural language prompts can be utilized to automate and streamline the complex backend API calls used to retrieve desired media content.

\section{Data}
The NFL’s Next Gen Stats (NGS) platform\footnote{\url{https://nextgenstats.nfl.com/}} is a comprehensive player and ball tracking system that captures and stores real-time data on every play from every NFL game since 2016. It stores large volumes of information, including information on players, teams, historical performances, and relevant advanced statistics. This data is captured through sensors in players’ pads and the ball. 

To build and test our agentic media search solution, our team used different statistical data such as team/player/play details, glossaries, question-answer pairs for media search, and backend OpenSearch
API call syntax from the NGS platform. Our team also worked closely with NFL Subject Matter Experts (SMEs) to prepare and validate the ground truths and to retrieve search results during solution development.

\subsection{Schema}
Due to highly granular nature of the NFL’s NGS data, the data is organized into different schemas (such as defense vs offense, passing vs rushing) to help with extracting stats and facilitating data organization. The NGS data is stored in OpenSearch and each schema consists of its own unique OpenSearch keys and fields. Table 1 \& 2 (Appendix) contains descriptions of key fields to demonstrate what the schemas look like. These schemas are passed into LLMs as context to help them understand the data types and definitions of each field.

\subsection{Question and Answer Pairs}
240 Question and Answer (QA) pairs were collected to represent user queries and their ground truth number of plays. The questions varied in complexity and were representative of questions that would be typically asked by NFL’s target end users such as research and production analysts - 30\% are easy questions that can be answered by straightforward API calls, 50\% are medium questions that require multiple filtering conditions and the remainder 20\% are complex questions that not only require deep football knowledge but also may involve multiple schema. 100 of the questions were set aside for the test set and the rest were used as development data to construct few-shots examples and enhance our prompting strategy. Examples of the QA pairs are shown in Table 3 (Appendix).

\section{Methodology}
We built an end‑to‑end, natural‑language “agentic” search experience - the system interprets a user’s query, asks clarifying questions when necessary, then composes and executes OpenSearch calls to fetch the matching plays and their media links. The workflow runs in AWS, is orchestrated with LangGraph
and is surfaced through a React front‑end (screenshots in the Appendix).  Figure~\ref{fig:agentic_workflow} gives the high‑level flow. Details of the design are described below.

\subsection{Agentic Workflow}
 The distinctly different schemas warrant the need for an agentic workflow that interprets which tables and fields are needed when constructing the API calls.

\begin{figure}
    \centering
    \includegraphics[width=0.75\linewidth]{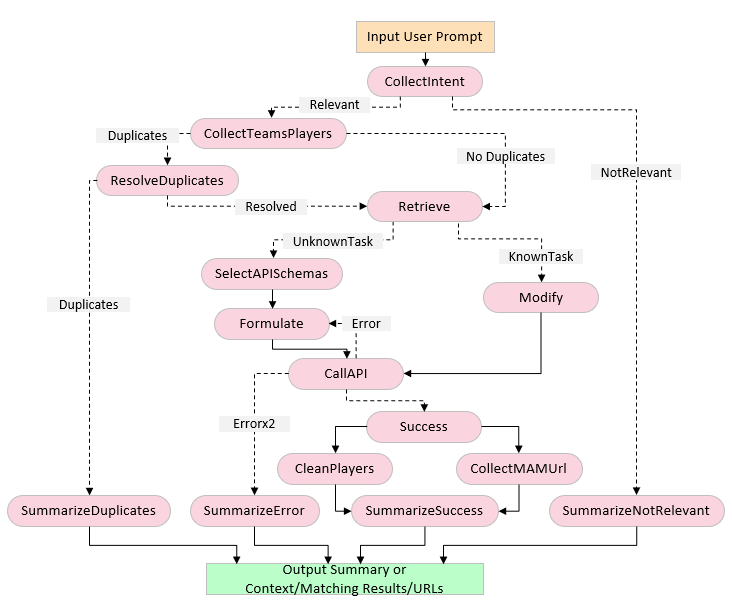}
    \caption{Overall diagram of the Agentic workflow}
    \label{fig:agentic_workflow}
\end{figure}

\noindent\textbf{Assess user intent}
A low‑latency model (such as Claude 3 Haiku) classifies the prompt.  Non‑football questions receive a polite rejection; valid football queries trigger the graph.  This guardrail prevents out‑of‑scope inputs from derailing later steps.

\noindent\textbf{Extract entities, actions, and conditions}
The LLM breaks down the prompt into Entities (players/teams), Actions (stats of interest), and Conditions.  
For a query such as “\emph{Find all plays where Patrick Mahomes throws a touchdown farther than 10 yards},” “Patrick Mahomes” is the Entity, “touchdown throw” is Action, “$>$10 yards” is the Condition. When situations with duplicate names appear (it's not uncommon for players in NFL to have same last names), the LLM prompts the user to confirm their player of interests and store confirmed player/team id for the session. 

\noindent\textbf{Select relevant API schema}: 
Based on extracted entities, actions and constraints, the system first conduct a semantic similarity search to identify if similar queries have been encountered in the past: 
\begin{itemize}
    \item If not, the system will pin down relevant context. A router LLM picks the most relevant NGS schema(s)—e.g.\ \textit{passing}, \textit{rushing}, \textit{team defense}, \textit{team offense}, etc. Each schema contains relevant fields that LLM can use to construct the API call. For example, the “\textit{passYards}” field corresponds to the yardage of passing, and “touchdown” is a boolean indicator of whether the play resulted in a touchdown. Only the relevant schemas are passed in as context to reduce overall context length and reduce the chance LLM picks inaccurate fields. 
    \item If similar queries are addressed before, the system will directly leverage the information instead of reformulating the problem. Details are described in \textbf{Semantic caching} below. 
\end{itemize}

\noindent\textbf{Formulate and execute API calls}: 
The API-formulation LLM receives team/ player IDs, reference schema, and few-shot demonstrations on how APIs are constructed. It reasons step-by-step, maps extracted actions and conditions to corresponding fields and emits the final API call with values assigned to each field. Figure 2 from Appendix shows a sample of formulated API call and reasoning LLM provides. A python runner executes the API call and an LLM summarizes the returned results into natural language responses to the user; Upon syntax failure, the LLM will attempt auto-correct using error messages. If the final API call still fails after three tries, the model will prompt the user to rephrase original query and run the workflow again.

\noindent\textbf{Conversational experience}: Follow‑up conversations inherit prior context. After “\textit{Find me all the plays where Patrick Mahomes throws a touchdown farther than 10 yards},” a user may ask “\textit{What about all the throws that were intercepted?}” The LLM implicitly keeps “Patrick Mahomes” as the entity, and will rephrase the user’s prompt to “\textit{Find all the plays where Patrick Mahomes throws are intercepted.}”

\noindent\textbf{Semantic caching}: 
We optimize latency and accuracy by storing prior queries and responses with player/team names redacted and replaced by placeholders (e.g.\ “[PLAYER]”, “[TEAM]”). When a new query gets submitted, the system first applies a vector search to find similar queries in the past. If a high score is achieved (meaning highly-similar queries have been addressed in the past), the cached API calls with placeholders will be replaced with the fresh IDs and executed directly, allowing the system to bypass expensive reformulation. 

\begin{figure}
    \centering
    \includegraphics[width=\linewidth]{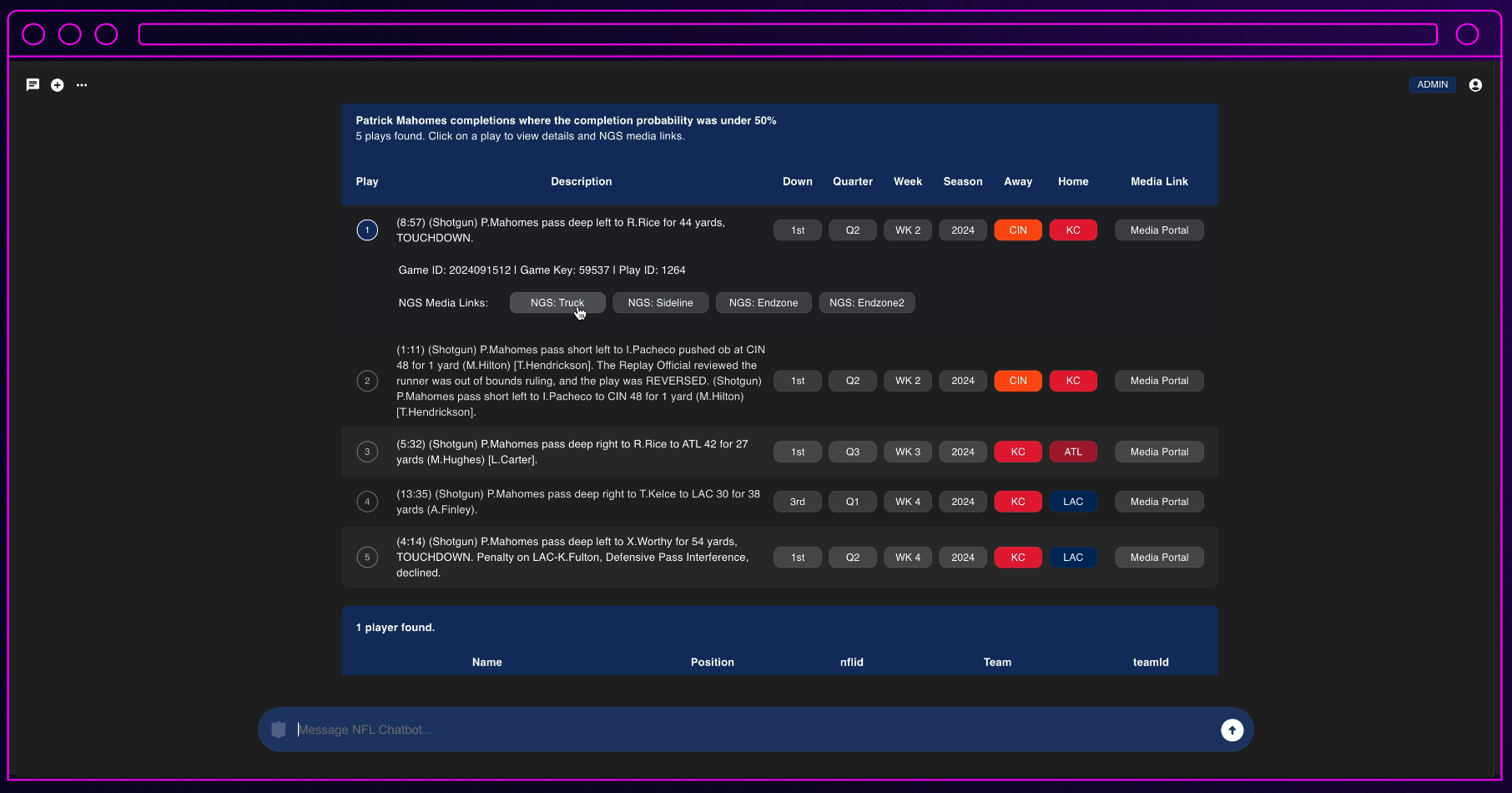}
    \caption{Links to retrieved assets embedded in React frontend}
    \label{fig:react_interface}
\end{figure}

\noindent\textbf{Link plays to assets in Media Asset Management(MAM) solution}: 
Finally, the play IDs are serialized and sent to NFL's MAM system, which will retrieve unique URLs for each asset. The React UI embeds the returned URLs. See Figure~\ref{fig:react_interface} for an example where a user can click on any of the returned NGS Media Links to access the media content directly in a MAM. At this point, the end-to-end content searching experience with natural language is complete.

\subsection{System Infrastructure}
The entire agentic workflow is implemented in the AWS ecosystem. There are 3 major components of this system. These are: 1) LLM model using Amazon Bedrock converse API, 2) Memory using Redis and 3) Workflow orchestration using LangGraph. The overall architecture diagram of the infrastructure is included in Appendix Figure 5.

\section{Results}
To develop the solution, our team used 100 QA pairs to refine our prompts and iteratively collected feedback from NFL SMEs to create a smooth user experience. The answer is considered accurate if the constructed APIs contain the correct filter\&values and the count of records matches with ground truth value. During testing, the solution achieved over 95\% accuracy on the 100 QA pairs. The solution has yielded notable process improvements across the NFL's media teams and 250 users were onboarded within the first month after its deployment. Users reported significant time savings: With free-form natural language queries, users can now search for and retrieve relevant video in an average of 30 seconds, down from over 10 minutes per query, giving them more time to focus on finding creative insights. They can also ask follow-up questions if they deem certain results returned interesting, so they can maintain the research flows without navigating between different tools.

\section{Conclusion}
In this paper, we discussed the implementation and efficacy of LLM-driven agentic workflows in transforming natural language queries into complex API calls within the NFL's Next Gen Stats system. Today, this solution is used daily by all of the NFL’s media editors. For simple searches that are expected to return videos for only a few specific plays, users can find relevant videos nearly 6 times faster than before. This time savings for finding relevant videos scales up to 20 times faster as search complexity increases and more game plays match the user’s prompt. In the future, there are opportunities to further enhance the capabilities of the tool and create fan-facing version of the solution, providing richer contents and offerings to hundreds of millions of avid football fans worldwide. 

%
%
%
\bibliographystyle{splncs04}
\bibliography{main}
\clearpage
\appendix




\section*{Appendix}

\section{Data}
\subsection{Schema}

\begin{table}[h!]
\caption{Sample keys from “rushing” schema. The schema contains name of the fields, valid values, types, and corresponding explanations of the fields. The information helps LLMs formulate accurate and executable API calls.}\label{tab1}
\begin{tabular}{|l|l|l|l|p{6cm}|}
\hline
\textbf{OpenSearch Key} & \textbf{Type} & \textbf{Format} & \textbf{Values} & \textbf{Explanation} \\ 
\hline
nflId & one-of-list & integer & & NFL ID is a unique identifier for an NFL player. Rusher ID is the NFL ID of the player who attempted a run on a play. \\ 
\hline
rushYards & range & float & (0, inf) & Rush Yards is the total distance in yards gained by a rusher on a run play or series of carries. \\ 
\hline
touchdown & singular & integer & [0, 1] & A rush attempt that results in a touchdown (6 points for the offense). \\ 
\hline
\end{tabular}
\end{table}

\begin{table}[h!]
\caption{Sample keys from “defense” schema. The schema contains name of the fields, valid values, types, and corresponding explanations of the fields. The information helps LLMs formulate accurate and executable API calls.}\label{tab1.2}

\begin{tabular}{|l|l|l|l|p{5cm}|}
\hline
\textbf{OpenSearch Key} & \textbf{Type} & \textbf{Format} & \textbf{Values} & \textbf{Explanation} \\ 
\hline
nflId & one-of-list & integer & & NFL ID is a unique identifier of an NFL player. Defender ID is the NFL ID of a player on the field for a defensive scrimmage play. \\ 
\hline
playerAlignmentEDGE & singular & integer & [0, 1] & The defender is aligned as an edge defender (aka defensive end). alignment = EDGE\\ 
\hline
alignmentDirection & one-of-list & string & [LEFT, RIGHT] & Alignment Direction is the side of the field the defender lines up on from the perspective of the defense (Left or Right). A player lined up on the left side of the defense is on the offense's right, while a player lined up on the right side of the defense is on the offense's left. \\ 
\hline
\end{tabular}
\end{table}

\clearpage
\subsection{Question and Answer Pairs}

\begin{table}[h!]
\caption{Sample queries and their ground truth play counts used for development and evaluation.}\label{tab2}
\begin{tabular}{|p{5cm}|p{2cm}|p{3cm}|p{4cm}|}
\hline
\centering \textbf{Query} & \textbf{Ground Truth Play Count} &\textbf{Complexity} & \textbf{Relevant Schema Fields} \\
\hline
How many passes did Patrick Mahomes throw during the 2022 regular season? & 648 & Easy & Patrick Mahomes 2022 Reg Play Type = Pass \\
\hline
How many touchdown passes greater than 10 yards did Patrick Mahomes throw during the 2022 regular season? & 12 & Medium & Patrick Mahomes 2022 Reg Play Type = Pass Pass TD | Pass Yards \textgreater 10 \\
\hline
How many touchdown passes greater than 10 yards did Patrick Mahomes throw during the 2022 regular season from Under Center? & 1 & Difficult & Patrick Mahomes 2022 Reg Play Type = Pass Pass TD | Pass Yards \textgreater 10 | Team Off Formation = Under Center \\
\hline
\end{tabular}
\end{table}

\clearpage
\section{System Interface}

\begin{figure}[h!]
\includegraphics[width=\textwidth]{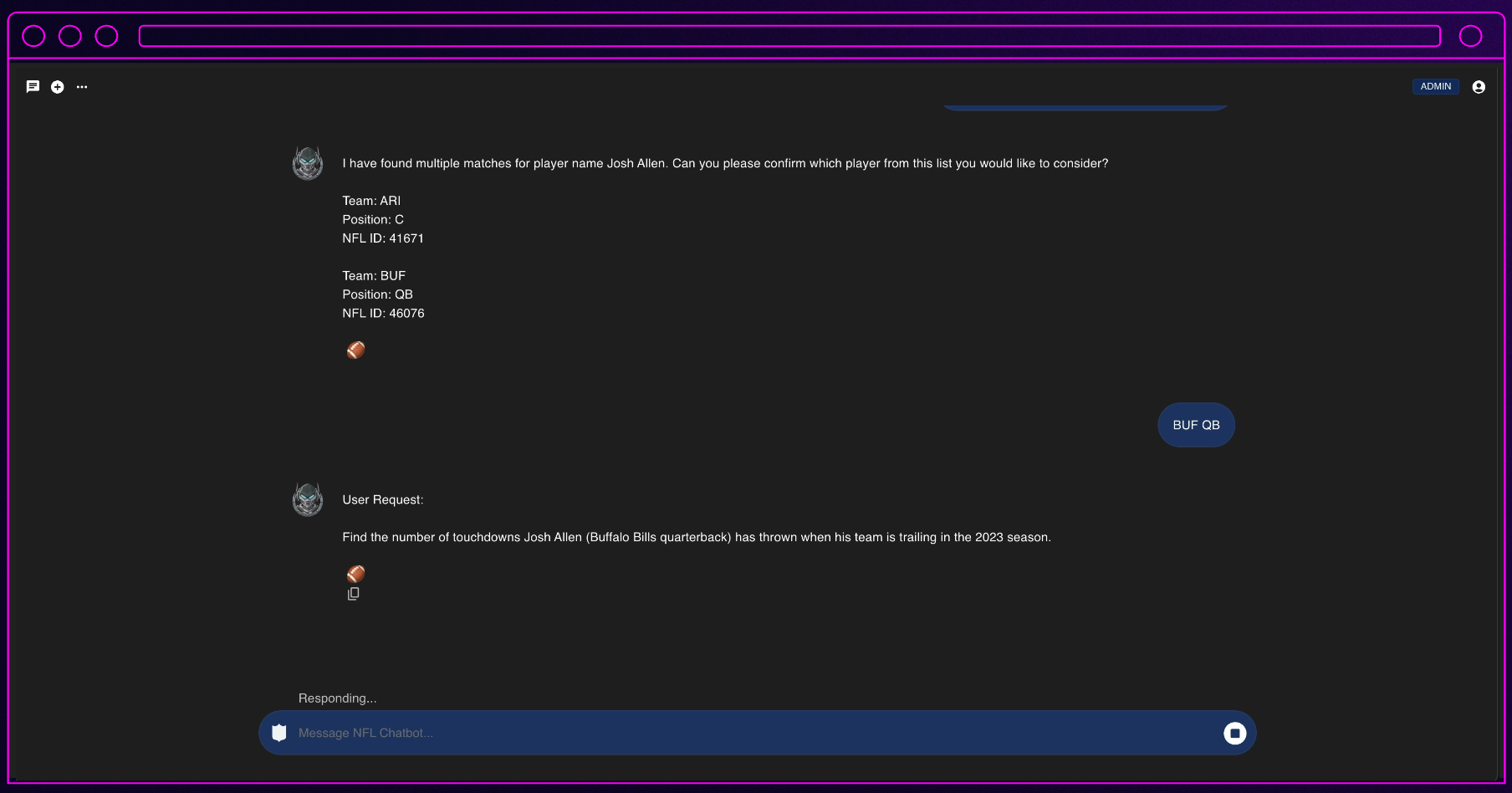}
\caption{The solution clarifies with the user when their initial prompt contains a player name that is tied to multiple distinct players in the database.} \label{}
\end{figure}

\begin{figure}[h!]
\includegraphics[width=\textwidth]{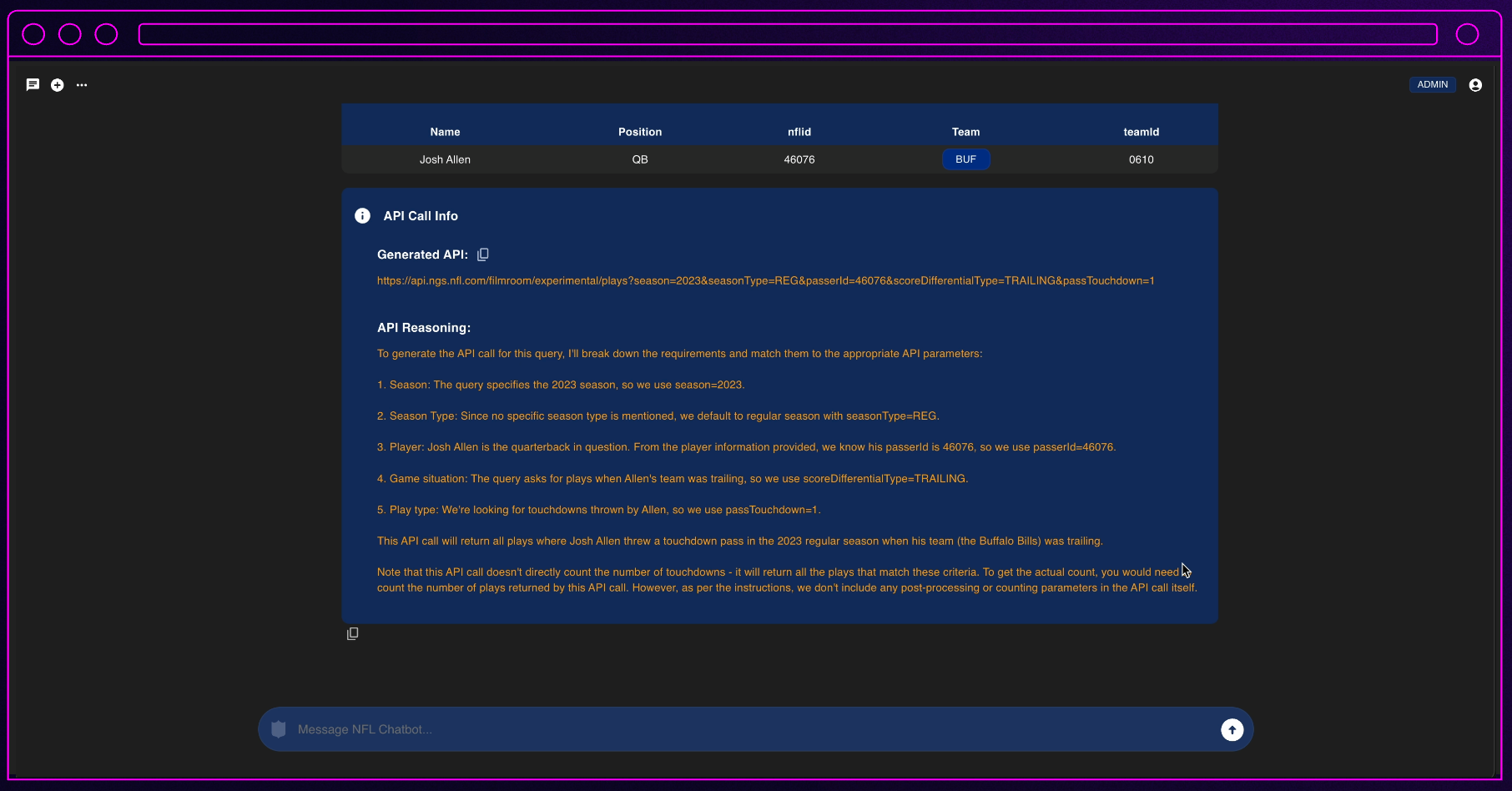}
\caption{Reasoning step and formulated API.} \label{}
\end{figure}

\begin{figure}[h!]
\includegraphics[width=\textwidth]{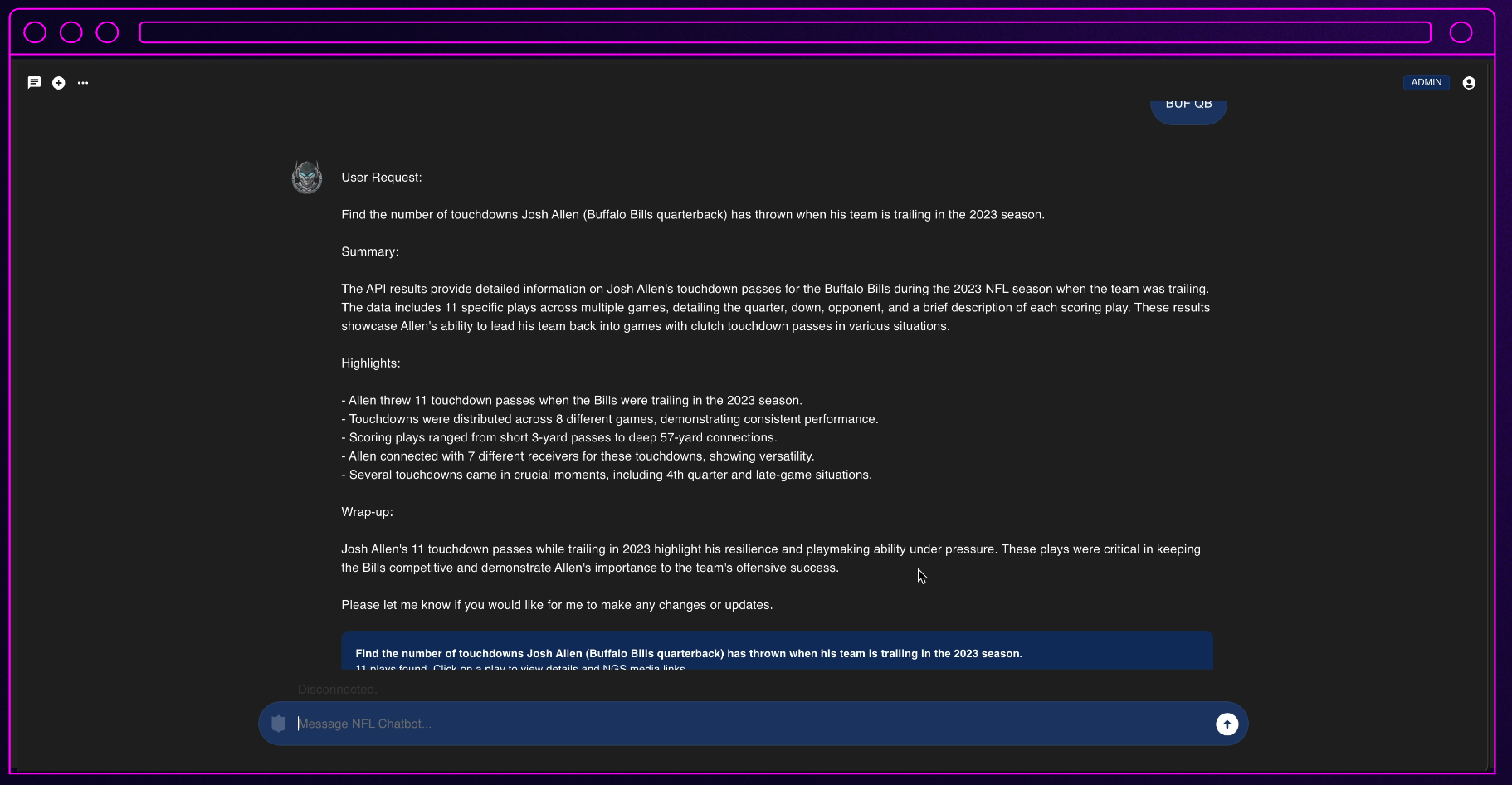}
\caption{Natural language response of the query.} \label{}
\end{figure}

\begin{figure}[h!]
\includegraphics[width=\textwidth]{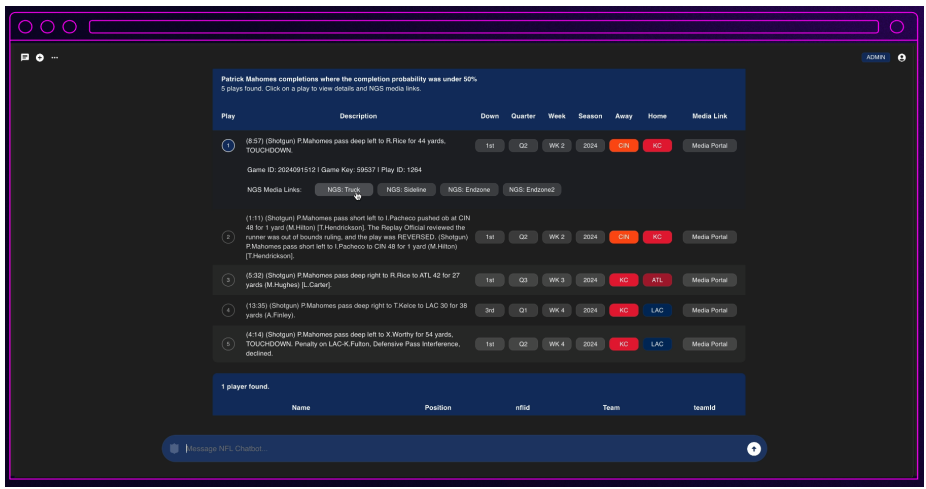}
\caption{Links to the retrieved plays’ associated media content within a Media Asset Management solution (MAM).} \label{}
\end{figure}

\clearpage
\section{System Architecture}

\begin{figure}[h!]
\includegraphics[scale=0.50]{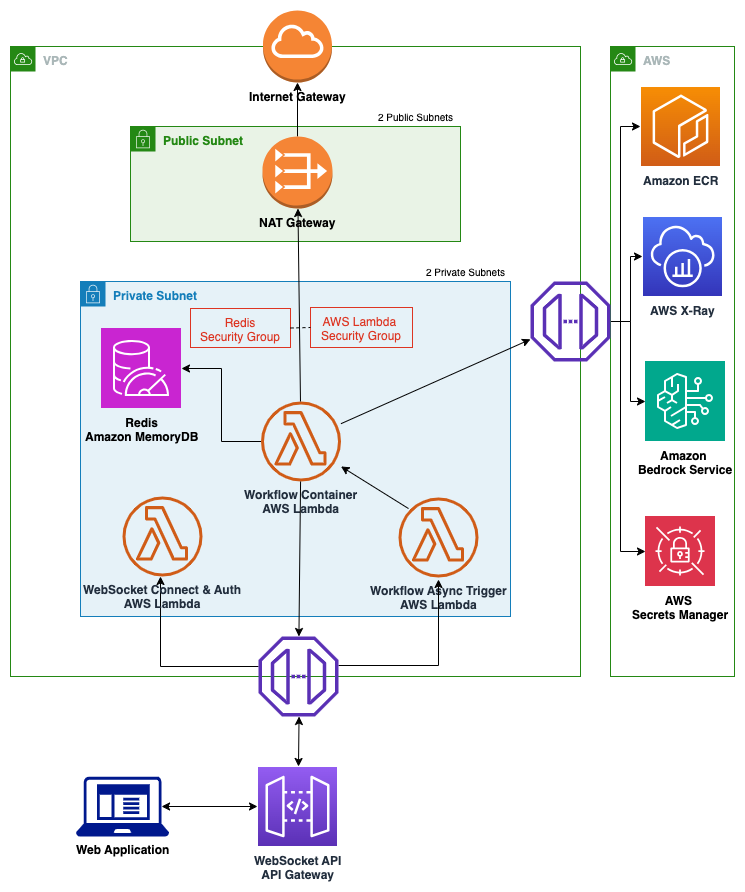}
\caption{Overall architecture of the infrastructure in AWS ecosystem.} \label{fig:infrastructure}
\end{figure}


\end{document}